# Can an AI System Be Creative?

## A Critical Perspective from Art and Engineering

**Ivan Magrin-Chagnolleau**

*Affiliated Scholar, Chapman University, Orange, California*

**Abstract**

This paper examines the question of whether artificial intelligence (AI) systems can be creative, approached from the dual perspective of a researcher trained in electrical engineering, pattern recognition, machine learning, and neural networks, who has also spent most of his life engaged in the arts as actor, stage and film director, writer, composer, and visual artist, and in philosophy. Drawing on Margaret Boden's foundational framework — both her three properties of creativity (novelty, surprise, and value) and her three types of creative processes (combinatorial, exploratory, and transformational) — the paper argues that AI systems are structurally incapable of creativity in its strongest sense. While they exhibit genuine capability in the domain of combinatorial creativity, they are significantly bounded in exploratory creativity, and fundamentally incapable of transformational creativity. The paper further argues that the most important limitation of current AI systems is not the absence of novelty per se, but the absence of any mechanism for serendipity, accident, or the unexpected — all of which play a central role in the phenomenology of creativity — and the absence of any subject position from which to recognize and welcome such chance events. The paper concludes by proposing a model of human-AI creative collaboration that is both realistic and generative, illustrated by several concrete experiments. The paper is itself a demonstration of the thesis it advances: it was composed through a deliberate human-AI collaborative process, which is described in the methodological note that opens it.

**Titre français**

# Un système d'IA peut-il faire preuve de créativité ?
## Une perspective critique issue de l'art et de l'ingénierie

**Résumé**

Cet article examine la question de savoir si les systèmes d'intelligence artificielle (IA) peuvent faire preuve de créativité, en adoptant la double perspective d'un chercheur formé en génie électrique, en reconnaissance de formes, en apprentissage automatique et en réseaux neuronaux, qui a également consacré la majeure partie de sa vie aux arts – en tant qu'acteur, metteur en scène de théâtre et réalisateur de cinéma, écrivain, compositeur et artiste visuel – ainsi qu'à la philosophie. S'appuyant sur le cadre théorique fondamental de Margaret Boden — à la fois ses trois propriétés de la créativité (nouveauté, surprise et valeur) et ses trois types de processus créatifs (combinatoire, exploratoire et transformationnel) — cet article soutient que les systèmes d'IA sont structurellement incapables de créativité au sens le plus fort du terme. S'ils font preuve d'une réelle capacité dans le domaine de la créativité combinatoire, ils sont considérablement limités en matière de créativité exploratoire et fondamentalement incapables de créativité transformationnelle. L'article soutient en outre que la limitation la plus importante des systèmes d'IA actuels n'est pas l'absence de nouveauté en soi, mais l'absence de tout mécanisme permettant la sérendipité, l'accident ou l'inattendu — qui jouent tous un rôle central dans la phénoménologie de la créativité — ainsi que l'absence de toute position subjective à partir de laquelle reconnaître et accueillir de tels événements fortuits. L'article conclut en proposant un modèle de collaboration créative entre l'humain et l'IA qui soit à la fois réaliste et génératif, illustré par plusieurs expériences concrètes. L'article est lui-même une démonstration de cette thèse.

# A Note on Method

This paper was composed through a deliberate human-AI collaborative process, which is itself a demonstration of the paper's central argument. The author began by dictating all thoughts and ideas in English using a voice transcription tool. The resulting transcription was then submitted to an AI system with the instruction to reorganize and refine it into a coherent academic draft. The same AI system was also asked to conduct a targeted survey of the relevant literature, which the author reviewed and validated. The AI system further assisted in generating several of the paper's illustrations, which the author evaluated, selected, and integrated. The author then revised the complete draft carefully for accuracy, nuance, register, and voice.

This process is not incidental to the paper's argument: it is a lived illustration of it. The AI system contributed to organization, drafting, literature review, and image generation. But every decision about structure, emphasis, argument, and expression was made — and validated — by the human author. The AI neither conceived the thesis nor evaluated whether the paper has merit. That responsibility, as this paper argues, remains irreducibly human.

The illustrations in Figures 1, 2, and 3 are the author's own photographs and artworks. The illustrations in Figure 4 were generated using the Ideogram AI image generation platform, directed and selected by the author. Figure 5 was generated by an AI image generation system and selected by the author. Their presence in the paper constitutes a dimension of Experiment 4: the AI generated; the author evaluated, selected, and integrated.

# 1. Introduction

The question of whether machines can be creative has been debated since long before artificial intelligence became a recognized field of research [Moruzzi 2025]. It has, if anything, become more urgent with the recent emergence of large language models and generative AI systems capable of producing text, images, music, and code at remarkable scale and speed. And yet urgency is not the same as clarity. The public discourse around AI creativity is often marked by either excessive enthusiasm — "AI is now creative!" — or reflexive dismissal — "AI is just a sophisticated statistical machine." Both positions, I will argue, miss something important. A more precise and useful answer requires a more precise and useful framework.

I approach this question from an unusual vantage point. I was trained as an electrical and computer engineer, with a doctoral focus on pattern recognition, machine learning, neural networks, and statistical models. At some point in my career, I decided to redirect my research toward the humanities, and in particular toward questions of creativity, artistic practice, and lived experience. But the computational background has never fully left me; it remains, as it were, a deep structure beneath the surface of my current work. Alongside this research trajectory, I have spent most of my life engaged in the arts — as an actor, stage and film director, writer, composer, and visual artist, and in philosophy. I have also given numerous talks on the topic of AI and creativity over the past several years, across academic and professional contexts. This paper is an attempt to bring these strands together into a more sustained and systematic argument.

The core claim I wish to advance is this: AI systems are, by definition and by construction, incapable of creativity in its strongest sense. This is not a vague philosophical intuition but a structural claim that follows from what AI systems actually are and how they actually work. At the same time, this claim requires important nuance. There is a spectrum of creativity, and AI systems are not equally incapable across all of it. To make this argument with the precision it requires, I rely on what remains the best available conceptual framework for thinking about creativity and computation: the work of Margaret Boden.

# 2. State of the Art

The scholarly literature on AI and creativity is extensive and growing rapidly. What follows is a focused overview of the contributions most directly relevant to the argument developed here.

The foundational work is that of **Margaret Boden** (1936–2025), Research Professor of Cognitive Science at the University of Sussex, who devoted much of her career to the intersection of artificial intelligence and creativity [British Academy 2025]. Her landmark 1998 paper "Creativity and Artificial Intelligence" (*Artificial Intelligence*, 103, 347–356) and her book *The Creative Mind: Myths and Mechanisms* (1990, revised 2004) established the conceptual vocabulary that still structures the field today. In her framework, creativity is defined by three properties — novelty, surprise, and value — and creative processes are classified into three distinct types: combinatorial, exploratory, and transformational. Boden herself noted that "AI will have less difficulty in modelling the generation of new ideas than in automating their evaluation" [Boden 1998] — a remark whose prescience has only grown more evident in the age of generative AI. Her passing in July 2025 marks the loss of one of the most rigorous and genuinely interdisciplinary thinkers the field has produced. It is fitting that this paper, written in the months following her death, takes her framework as its central point of reference.

Recent scholarship has revisited Boden's framework in the light of contemporary AI systems. **Moruzzi (2025)**, in a comprehensive review published in *Philosophy Compass*, traces the history of AI's successive attempts to replicate each of Boden's three properties of creativity, and concludes with a reflection on how the third property — value — raises societal challenges that exceed and outlast the question "can AI be creative?" [Moruzzi 2025]. **Ismayilzada, Paul, Bosselut, and van der Plas (2024)**, in a widely cited survey ("Creativity in AI: Progresses and Challenges," arXiv:2410.17218), find that while current AI models are largely capable of producing linguistically and artistically plausible outputs — poems, images, musical pieces — they struggle significantly with tasks requiring creative problem-solving, abstract thinking, and compositionality, and their outputs suffer from a "lack of diversity, originality, long-range incoherence and hallucinations" [Ismayilzada et al. 2024]. **Schapiro, Black, and Varshney (2025)** have formalized Boden's notion of transformational creativity using directed acyclic graphs, proving that modifications to the axioms of a conceptual space carry the highest transformative potential — precisely the kind of operation that current AI systems are constitutively incapable of performing [Schapiro et al. 2025].

On the philosophical and phenomenological side, **Lockhart (2024)** examines the limitations of large language models through the lens of embodied cognition, arguing — drawing on Rollo May and others — that the deeply personal, relational, and accidental dimensions of human creativity remain beyond the reach of AI [Lockhart 2024]. **Bianchi, Branchini, Uricchio, and Bongelli (2025)**, working at the intersection of cognitive psychology and empirical aesthetics, highlight the multidimensionality of creativity and call for new research frameworks adequate to the question of value in AI-generated artworks [Bianchi et al. 2025].

I have also addressed adjacent questions in two prior publications that are directly relevant to this paper. The first examines the intervention of chance in the artistic creative process from a phenomenological perspective [Magrin-Chagnolleau 2021]; the argument developed in Section 4.4 below draws extensively on that work. The second explores phenomenological and enactive approaches in artistic creation and pedagogy [Magrin-Chagnolleau 2025]. A third publication, co-authored with colleagues at the PRISM laboratory, documents a concrete experiment in computer-assisted musical composition that provides one of the illustrative cases in Section 6 [Ariani et al. 2023].

This paper builds on these contributions, but its distinctive contribution lies in combining a structural argument — rooted in the technical architecture of AI systems — with a phenomenological argument — rooted in the lived experience of creative practice — and in offering a graduated analysis of AI's capabilities and limitations across all three of Boden's creative types. It is also, to my knowledge, one of the few papers to place the role of accident and serendipity at the center of the argument, and to connect that argument to the question of the subject who perceives and welcomes chance.

# 3. Defining Creativity: Boden's Framework

## 3.1 The Three Properties of Creativity

Boden's definition of creativity is, in her own formulation, deceptively simple: "Creativity is the ability to come up with ideas or artefacts that are new, surprising and valuable" [Boden 2004]. Each of these three properties deserves careful attention, because each will serve as a test for AI systems in the analysis that follows.

**Novelty** refers to the requirement that a creative product be genuinely original — not merely a recombination of what already exists, but something that introduces a new element into the world. Boden distinguishes between *psychological* novelty (new to the individual) and *historical* novelty (new to the world as a whole). For the purposes of this paper, we are concerned primarily with the latter: the kind of novelty that changes the landscape of what is possible.

**Surprise** is the property that makes a creative product unexpected — it should not be what we would have predicted, even in retrospect. Surprise is what distinguishes a truly creative act from a merely competent one. A technically correct haiku that says exactly what the form leads us to expect is not surprising; a genuinely creative haiku reframes how we see something, producing a shift in perception that no amount of prediction could have prepared us for.

**Value** is perhaps the most demanding of the three properties, and the most philosophically complex. A creative product must not merely be new and surprising — it must be *worth* something. Value, in this context, is aesthetic, intellectual, emotional, or practical, depending on the domain. Crucially, value is not an intrinsic property of the object itself: it requires an evaluating subject. There is no value without a witness, and the identity and expertise of that witness matters enormously.

## 3.2 The Three Types of Creative Processes

In addition to characterizing what creativity *is*, Boden also characterizes how it *happens*. She identifies three distinct types of creative processes [Boden 2004], each corresponding to a different relationship between the creative act and the existing conceptual landscape.

**Combinatorial creativity** involves the production of novel combinations of familiar ideas. Much of what we recognize as creativity in everyday life — metaphors, analogies, stylistic fusions, surprising juxtapositions — falls into this category. The surprise it produces comes from the unexpectedness of the combination, not from the introduction of genuinely new conceptual elements. The materials are pre-existing; what is new is the way they are brought together.

**Exploratory creativity** involves the systematic exploration of an existing conceptual space — pushing to its edges, testing its limits, discovering what is possible within its established structure. A jazz musician improvising within a harmonic framework, or a poet exhausting every possibility of the sonnet form, is engaged in exploratory creativity. The creative act consists in discovering what a space contains, including what it contains at its extremities — and crucially, in recognizing the edges of that space when they are reached.

**Transformational creativity** is the most radical and the rarest form. It involves restructuring the conceptual space itself — changing the fundamental rules in a way that makes previously impossible ideas possible. Cubism, twelve-tone music, non-Euclidean geometry, the theory of relativity: these are all instances of transformational creativity. What is distinctive about transformational creativity is that the new conceptual space cannot be reached by any amount of exploration within the old one. It requires a break, not a progression.

This taxonomy will serve as the organizing structure of the analysis in Section 5.

# 4. The Structural Limitations of AI Systems

Before analyzing AI's performance across Boden's three creative types, it is necessary to establish clearly what AI systems actually are. I use the term "AI system" deliberately, rather than "AI model" or "AI algorithm," to acknowledge the complexity of contemporary deployments: what a user interacts with today is typically not a single model but an ensemble of components — large language models, retrieval systems, ranking algorithms, safety filters, user interface layers — working in concert. This architectural complexity does not change the fundamental argument, but it is worth bearing in mind that "AI" in practice refers to a sociotechnical system, not a single computational entity.

## 4.1 The Data-Bound Problem: Novelty

An AI system, by definition and by construction, relies on a large dataset of pre-existing human-generated (and sometimes AI-generated) content. Its outputs are produced by identifying statistical patterns in that dataset and generating responses consistent with those patterns. This means, structurally, that everything an AI system produces already exists — in some form — within its training data. It cannot produce something that has no precedent anywhere in its data, because its generative process is a function of that data. It does not observe the world and form new concepts; it processes representations of the world that human beings have already produced and identifies statistical regularities among them.

This is not a contingent limitation that could be overcome by a better algorithm or a larger dataset. It is a constitutive feature of how these systems work. Historical novelty — the introduction of something genuinely new into the world — is therefore structurally unavailable to an AI system.

## 4.2 The Statistical Convergence Problem: Surprise

AI systems are, at their core, probability engines. When generating an output, the system calculates, for each possible next element — token, pixel, note — the probability of that element given the context, and selects from the high-probability region of that distribution. Even when parameters are adjusted to introduce variation, the system is operating within a probabilistic framework shaped by what has most frequently appeared in its training data. The output that maximizes probability is, by definition, the most *average* output — the one most consistent with the existing corpus.

This has a direct and important consequence for surprise. Genuine surprise, in Boden's sense, requires that the output not be what we would have predicted. But an AI system trained to produce statistically probable outputs is, by design, producing exactly what the data predicts. It can produce variation, but the variation is sampled from a distribution of the already-seen. The structural tendency is toward the average, not toward the surprising.

## 4.3 The Evaluation Problem: Value

Perhaps the deepest limitation of current AI systems — and the one most often underestimated — is their inability to evaluate the value of their own outputs. An AI system can compare its output to patterns in its training data and assess whether the output is consistent with highly rated or frequently selected examples. But this is not aesthetic judgment. It cannot ask: "Is this beautiful? Is this moving? Is this worth something?" It can only ask: "Is this statistically consistent with what human beings have previously called beautiful, moving, or valuable?"

The distinction matters enormously. The history of art is full of works that were initially rejected — by critics, by audiences, by institutions — and later recognized as masterpieces. Conversely, it is full of works celebrated in their moment and then forgotten. The capacity to recognize genuine value, especially at the moment of creation and before any external validation, is one of the most distinctively human aspects of the creative process. An AI system trained on historical evaluations cannot access this: it can only reproduce the evaluative patterns of the past.

As Boden herself noted in 1998 — and as has become even more evident in the years since — AI's difficulty in automating the *evaluation* of creativity is more profound than its difficulty in *generating* creative outputs [Boden 1998]. Generating a plausible poem is a statistical achievement. Recognizing that the poem is genuinely great is something else entirely.

## 4.4 The Phenomenological Dimension: Accident, Chance, and the Unexpected

There is a fourth limitation that does not appear explicitly in Boden's framework but that my own research — and my own experience as a practicing artist — has convinced me is central to any serious account of creativity: the role of accident, serendipity, and the unexpected.

I have explored this dimension at length in a previous publication [Magrin-Chagnolleau 2021], which examines the intervention of chance in the artistic creative process from a phenomenological standpoint — that is, from the point of view of the lived experience [Magrin-Chagnolleau 2025]. The argument I develop here is closely connected to that earlier work, and I refer the reader to it for a more detailed treatment.

A semantic exploration of the word 'chance' itself is instructive. In philosophy and science, chance is defined as "that which falls under the laws of probability and is not deliberate" — it is the non-deliberate, the non-foreseen, the unplanned. Henri Bergson and Paul Valéry both insist, crucially, that there is no chance without a subject: chance is always *for* someone, perceived and recognized by a consciousness that is affected by it. As Bergson writes, "chance is therefore a mechanism, behaving as if it had an intention" — and Valéry makes clear that to attribute something to chance is always to introduce a human subject who is "particularly sensitive to this trait" [Magrin-Chagnolleau 2021]. This is a point of great consequence to which we will return.

Three examples from my own artistic practice may serve to ground this abstract argument. The first: visiting Muir Woods National Monument in California, I began photographing the giant redwood trees in a conventional way, then — at a moment I cannot precisely identify as a deliberate decision — began moving my camera during the exposure. Whether the first such image was taken by accident or arose from an intuition in the present moment, it was unpremeditated and unforeseen, and it opened an entire photographic series [Magrin-Chagnolleau 2021].

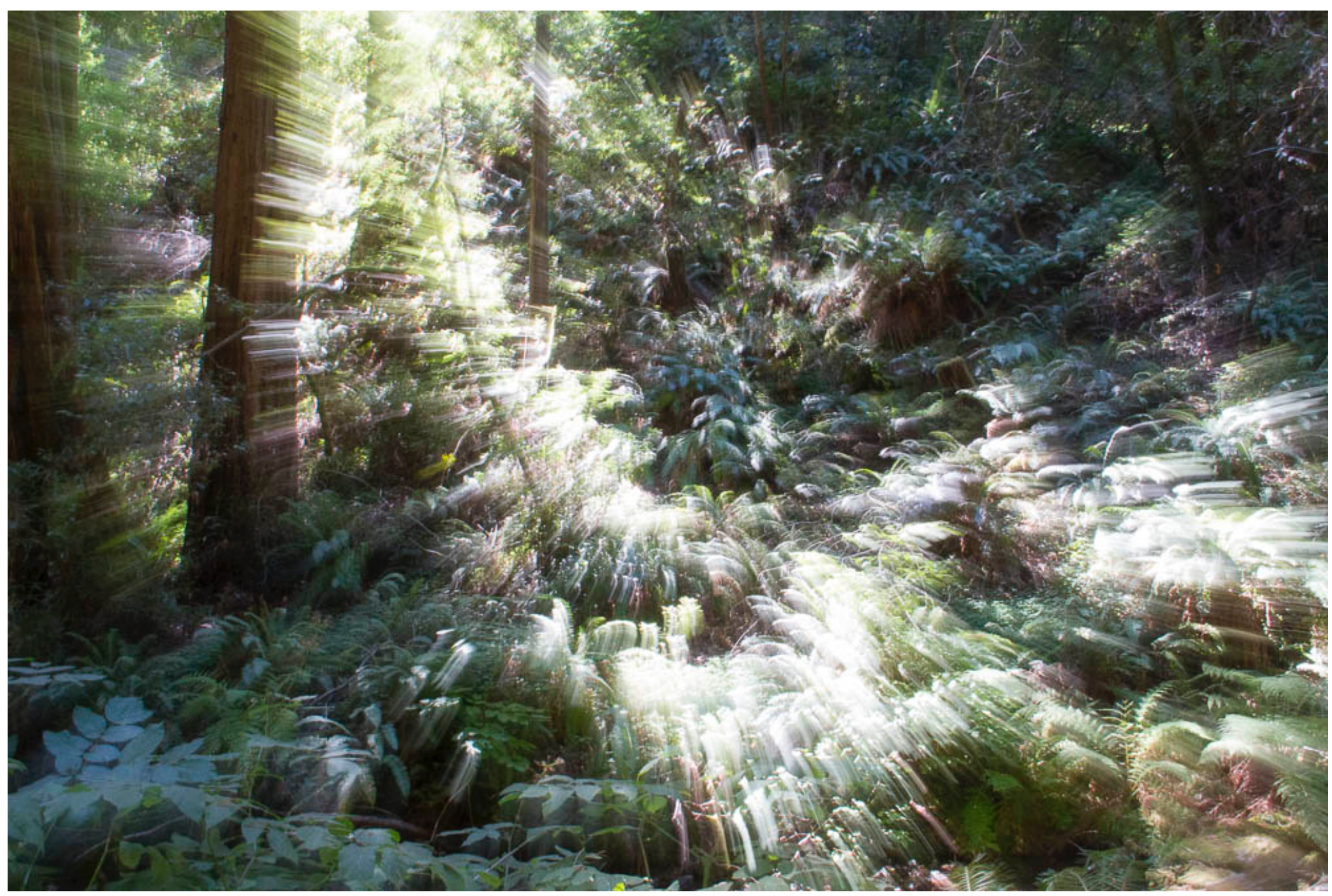

Figure 1: Muir Woods Spirits #26

The second: on a regional train in Auvergne, I took out my phone to photograph a beautiful landscape, and found in the resulting image a reflection of my own hands and phone in the train window. I had not planned this. The physical phenomenon that produced it is perfectly explicable; but I had not foreseen it, and it became the founding accident of an entire photographic series [Magrin-Chagnolleau 2021].

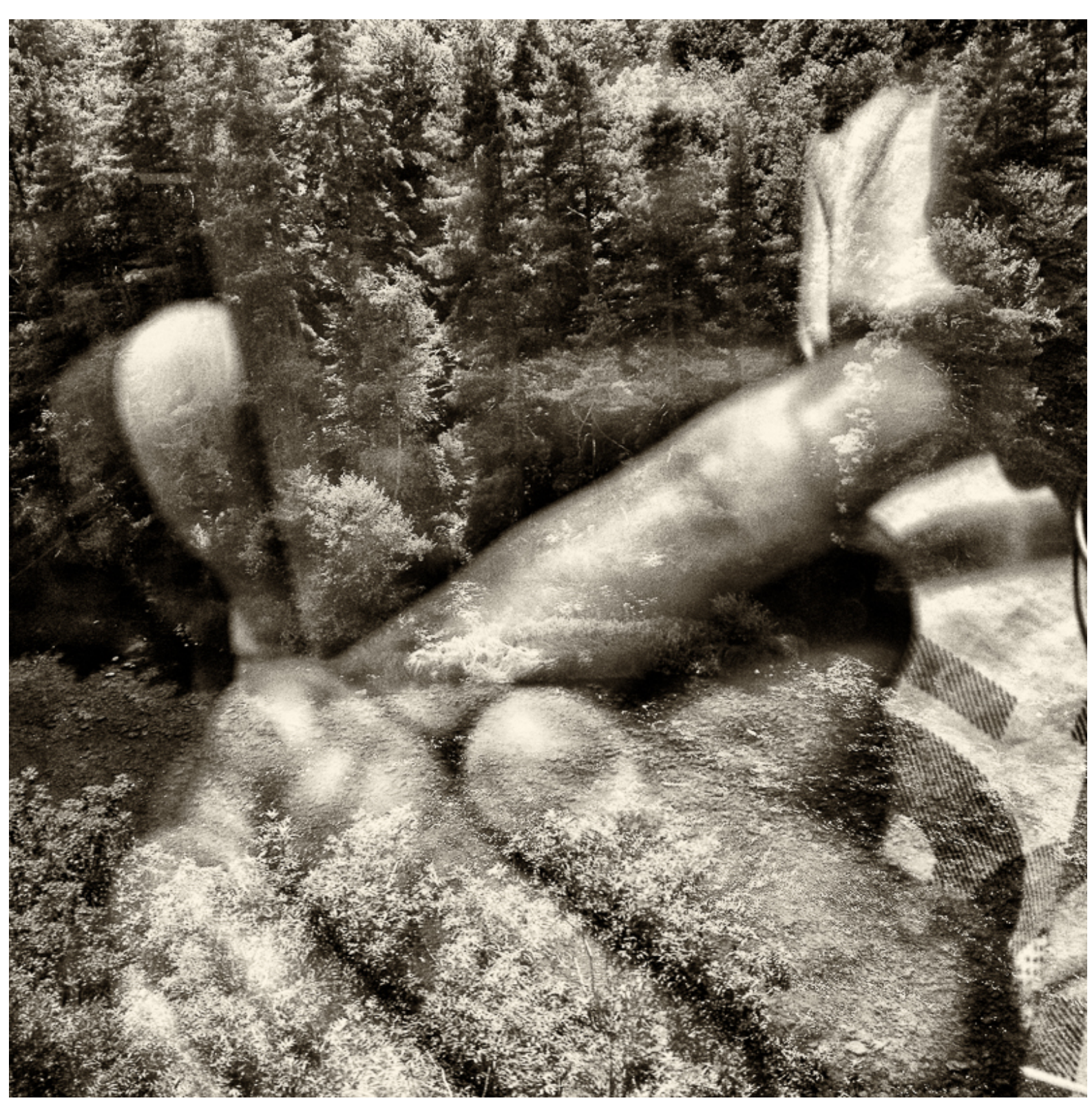

Figure 2: Handscape #31

The third, perhaps the most extreme: six beginner watercolors stored in a cardboard folder in an attic for fifteen years. When I found them again, humidity and capillarity had "composed" an entirely new watercolor on one of the blank sheets — a kind of meditative landscape that I could never have planned or predicted [Magrin-Chagnolleau 2021]. From the point of view of the lived experience, this was genuinely chance. It was also genuinely creative.

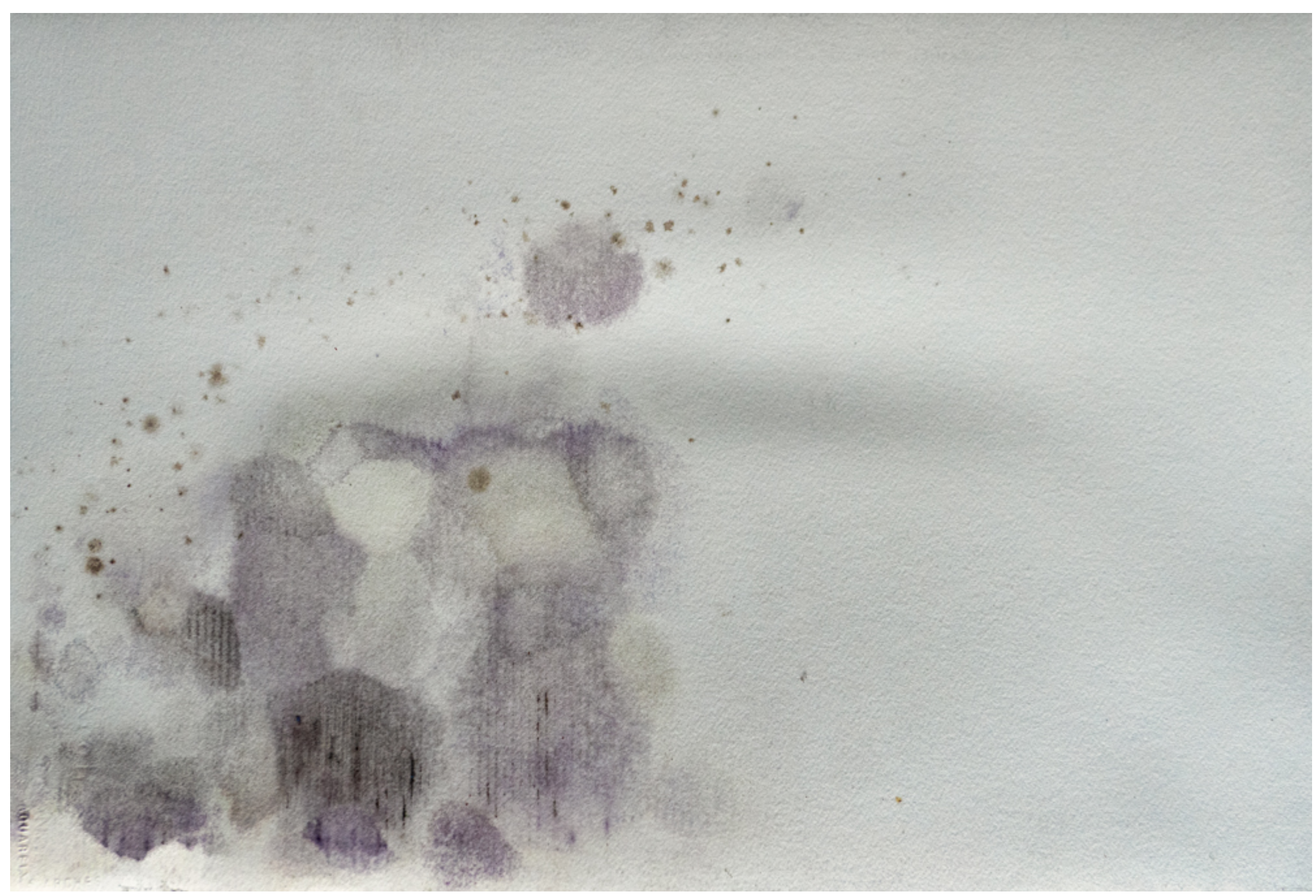

Figure 3: Sedimented Landscape

These examples illustrate a principle that runs through the literature on creativity and chance: *chance must be provoked*. As Romain Rolland writes, "chance always knows how to find those who know how to use it" — and Albert Camus speaks of "giving opportunities to this chance which, too often, is disturbed only when provoked" [Magrin-Chagnolleau 2021]. This is not a passive observation. It means that genuine creative practice involves cultivating a state of openness and readiness — a disposition to recognize the unexpected as an opportunity rather than an error, to welcome the accident rather than correct it. Jacques Monod, in *Le Hasard et la Nécessité*, went further still: "chance alone is at the source of all novelty, of all creation within the biosphere" [Monod 1970]. Whatever the current status of this claim after the advances of epigenetics, it points to something important: novelty, in the deepest sense, may be inseparable from the unpredictable.

All of this is structurally absent from AI systems. Everything in an AI system is, in a fundamental sense, *expected*: every output is a function of a deterministic or stochastically bounded process applied to training data. There is no mechanism for genuine surprise from within — no equivalent of the camera that moves unexpectedly, the reflection that appears in the window, the humidity that composes over fifteen years. Random noise can be introduced into generative processes, but introduced randomness is not genuine accident. It is controlled unpredictability — and controlled unpredictability is, paradoxically, a form of expectedness.

Moreover, and this is the compounding deficit: even if noise were somehow equivalent to accident, an AI system would have no subject position from which to *recognize* the accident as significant — no Bergson-Valéry subjectivity from which chance could be perceived as chance. An AI system cannot notice that something unexpected has happened and decide to follow it. It has no state of readiness, no capacity for the kind of present-moment attention that transforms an accident into a creative discovery. The two deficits — the absence of genuine accident and the absence of a subject to receive it — reinforce each other completely.

# 5. AI and the Three Types of Creativity

With these structural limitations established, we can now return to Boden's taxonomy and ask, for each type of creative process, what AI systems can and cannot do. The answer, as I noted at the outset, is graduated rather than binary.

## 5.1 Combinatorial Creativity: Where AI Has Genuine Foothold

Combinatorial creativity — the production of novel combinations of familiar ideas — is the domain where AI systems perform best, and where their performance is most genuinely impressive. Because they have access to vast amounts of data across many domains, AI systems can identify connections and combinations that a single human mind, with its necessarily limited scope of exposure, might never arrive at independently. A researcher working on a problem in molecular biology might never encounter a relevant metaphor from music theory; an AI system trained on both might make that connection effortlessly.

This capability is real, and it should not be minimized. In many practical creative contexts — brainstorming, ideation, the early stages of a project — the ability to rapidly generate unexpected combinations can be enormously generative. The experiments I describe in Section 6 are, at their core, instances of combinatorial creativity being put to productive use.

The critical limitation remains, however: the system cannot evaluate the value of the combinations it produces. It can generate, but it cannot judge. That distinction — between generation and judgment — is the fault line that runs through every discussion of AI creativity.

## 5.2 Exploratory Creativity: A Bounded Exploration

Exploratory creativity — the systematic traversal of an existing conceptual space — is a more nuanced and underanalyzed case. In one sense, AI systems are well-suited to exploration: they can rapidly generate many variants of a given structure, style, or form, efficiently covering a wide range of possibilities within a defined set of constraints.

But there is a crucial limitation that distinguishes AI exploration from genuine exploratory creativity. Genuine exploratory creativity involves not only traversing a conceptual space but also *recognizing its edges* — understanding where the space ends and what might lie beyond it. This recognition is often what leads to the most interesting discoveries: the artist who pushes a form to its limits, discovers what cannot be done within those limits, and thereby glimpses what might be possible beyond them. This is the moment where exploratory creativity becomes the precondition for transformational creativity.

AI systems cannot reliably perform this second act. They can generate outputs that vary within a space, but they have no meaningful vantage point from which to recognize the boundary of that space, let alone understand what it would mean to approach or exceed it. Their exploration is bounded by the limits of their training data, and they have no way of knowing that they are at a boundary rather than at the center. There is no sense of edge, no awareness of the frontier — which means there is no possibility of the creative leap that frontier-recognition makes possible. Furthermore, AI systems cannot evaluate the relative *interest* of different positions within the conceptual space they are exploring: they can produce many variants, but they cannot tell us which variant is most revealing, most surprising, most alive.

## 5.3 Transformational Creativity: The Definitive Limit

Transformational creativity — the restructuring of the conceptual space itself — is where AI's limitations are most absolute, and where the argument developed in Section 4 converges most forcefully.

Transformational creativity, by definition, produces something that could not have been produced by any process operating within the existing conceptual framework. It requires stepping *outside* the current paradigm and seeing it from a position that does not yet exist within it. This is precisely what AI systems cannot do. An AI system is, in the most fundamental sense, a product of its training data. Its conceptual space is defined and bounded by that data. It cannot step outside that space, because it has no position from which to do so. In Boden's terms, it cannot change the enabling constraints of the domain — it can only operate within them [Boden 2004].

Recent formal work confirms this intuition with mathematical precision. Schapiro, Black, and Varshney (2025) have shown, using a graph-theoretic model of conceptual spaces, that transformational creativity requires modifications to the axioms of the space — the most fundamental level of its structure. These are precisely the modifications that a statistical model trained to reproduce existing patterns is least capable of making [Schapiro et al. 2025].

Historically, all genuinely transformational creative acts have involved a human being willing to break with existing conventions in a way that could not be justified by those conventions at the time. Cubism did not emerge from an exploration of the space of existing painting styles; it emerged from a decision to abandon one of the most fundamental assumptions of Western pictorial representation — the single viewpoint. This decision was not statistically predictable from any corpus of existing paintings. It required a human vision — and, very likely, an accident of perception, an unexpected moment of seeing that could not have been planned. The role of the unexpected, of the accident, is once again at the center of the most creative act.

# 6. AI as Creative Collaborator

The analysis in the preceding sections might appear to suggest that AI systems are, for creative purposes, essentially useless. That would be an overcorrection. The argument is not that AI systems have no role in creative processes; it is that their role is that of a *collaborator*, not a *creator* — and a collaborator of a specific and limited kind. When used by someone with the expertise to guide the collaboration and evaluate its products, AI systems can be genuinely valuable creative partners.

## 6.1 Four Experiments

Let me describe four experiments I have conducted or been party to, which illustrate both the promise and the limits of AI as a creative collaborator.

**Experiment 1: Haiku composition.** In the autumn of 2025, I conducted a multi-phase experiment in which I asked an AI system to compose haikus under progressively more specific conditions. In the first phase, I simply asked for ten haikus, with no constraints on topic or style. In subsequent phases, I asked for haikus on specific topics (current news), haikus using three specific words (*consciousness*, *creativity*, and *free will*), interdisciplinary haikus combining two fields of knowledge, the system's ten "best, most mindblowing" haikus, and finally haikus on the theme of teaching. In total, the system produced sixty haikus across six batches.

Several observations emerged from examining the results. The formal structure — the 5-7-5 syllable pattern — was respected throughout, as one would expect: the constraint is highly specific, and the training data for haiku is vast. This demonstrates the system's genuine capability at the level of form, which is a real instance of combinatorial creativity: the AI correctly identifies and reproduces the structural rules of the genre.

More instructive, however, was what happened under constraint. When asked to use the three words *consciousness*, *creativity*, and *free will* in each haiku, the system converged almost immediately on a single template and applied it ten times with different nouns substituted in: *"Consciousness in [X] / Creativity in [Y] / Free will in [Z]."* The result was not ten haikus but one structure repeated ten times — *"Consciousness in trees / Creativity in leaves / Free will in the breeze"*; *"Consciousness in snow / Creativity in frost / Free will in* *the glow."* This is statistical convergence made visible: faced with a tight constraint, the system finds the most probable solution and reproduces it, with minimal variation.

Equally revealing was the "mindblowing" prompt. Asked to produce its ten best and most mindblowing haikus, the system produced the same kind of haiku it had always produced — slightly more abstract in vocabulary (*"quantum," "fractal," "parallel lives"*), but structurally and aesthetically indistinguishable from its earlier outputs. The reason is clear: the system has no capacity for aesthetic self-evaluation. It cannot know which of its haikus are better than others, because "better" requires a judgment of value that no statistical model can perform. Asked to produce its best work, it produces its most average work — because the most probable output *is*, by definition, the most average one.

A final detail worth noting: the system gave every haiku a title — *"Morning Dew," "Desert Mirage," "Eternal Echo."* Classical Japanese haiku do not have titles; the form is complete in itself. The AI added titles because the most statistically common Western representations of haiku in its training data include them. It reproduced the genre's most common form, not its most authentic or most sophisticated one.

None of the sixty haikus produced was, in my judgment, a great haiku — one that produces a sudden shift in perception, a genuine surprise, a moment of seeing something differently. Several were competent and occasionally pleasant. In one or two cases, a combination of images prompted me to think of a haiku I might write myself — which is the productive collaboration the paper describes. But the evaluative act — selecting what had value, noticing the formulaic convergence, recognizing the distance between formal correctness and genuine creativity — was performed entirely by the human reader. The AI had no access to any of it.

**1. To illustrate formal competence** (from batch 1 — nature haikus):

*Crimson leaves descend*
*Whispering secrets of fall*
*Earth's warm, final blend*

Competent, pleasant, unsurprising — the average of the corpus.

**2. To illustrate the template problem** (from batch 3 — consciousness/creativity/free will):

*Consciousness in trees*
*Creativity in leaves*
*Free will in the breeze*

*Consciousness in waves*
*Creativity in tides*
*Free will in the caves*

The clearest illustration of statistical convergence.

**3. To illustrate the failure of the "mindblowing" prompt** (from batch 5):

*Stars in endless night*
*Universe in boundless flight*
*Minds expand in light*

Aspires to profundity; achieves the average. This is the most concise demonstration that the system cannot evaluate its own output.

**Experiment 2: Autobiography in the style of French literary masters.** In the spring of 2026, I wrote the opening paragraph of what might be my autobiography — in French, my native language — and submitted it to ChatGPT (OpenAI; exact version unrecorded, for reasons discussed below) with the instruction to rewrite it successively in the style of Flaubert, Maupassant, Proust, Victor Hugo, and Balzac, then in a synthesis of all five, and finally in a modern but literary version of that synthesis. The original paragraph reads as follows:

> *Je suis né à Paris. En fait, je suis né à Montmartre, ce qui aujourd'hui n'est plus possible car la clinique où je suis né, la clinique Ornano rue Championnet, n'existe plus. La seule façon de naître encore à Montmartre est de naître à la maison, ce qui est de plus en plus rare. Je suis né à Paris, et pourtant mes parents n'y habitaient pas quand je suis né. Ils habitaient en région parisienne. Je ne sais pas pourquoi j'y suis né. Mais je porte cette naissance comme une sorte de titre honorifique.*

The results were, in several respects, remarkable. The system had access to extensive corpora of all five authors and was adept at identifying and reproducing their most prominent stylistic features. The Flaubert version correctly deploys his characteristic precision and sobriety — *"Ce lieu de naissance, désormais inaccessible, confère à mon arrivée dans le monde une singularité presque ironique"* — a sentence Flaubert might plausibly have written. The Proust version is the strongest output of the experiment: the long, folding subordinate clauses, the meditative relationship to time and memory, the way the sentence returns upon itself — *"ayant disparu depuis longtemps, comme tant de lieux qui furent autrefois les témoins silencieux de nos premiers instants"* — capture something genuinely Proustian, and this deserves honest acknowledgment. The modern literary synthesis, the seventh and final version, produces a spare, rhythmically controlled paragraph with a genuinely literary quality: *"Pour naître là-haut, il faudrait désormais le faire chez soi — et qui choisit encore cela ?"*

And yet several observations qualify this apparent success considerably. First, the Hugo version is the most revealing precisely because it is the most formulaic. It immediately deploys exclamation marks, apostrophes, and grand rhetorical declarations — *"Paris ! Paris, la ville lumière, la cité des révolutions et des rêves !"* — because these are the most statistically prominent and most frequently caricatured features of Hugo's style in any large training corpus. But Hugo's greatness does not reside in his exclamation marks; it resides in the unexpected turn where the grandiose suddenly collapses into the intimate, or the epic into the ironic. That dimension — the surprising move, the element that makes a reader stop — is entirely absent. The AI reproduces the most recognizable surface of Hugo's style, not its depth. This is combinatorial creativity at its most visible limit: the system combines the most statistically salient elements, and the result is closer to a stylistic pastiche than to a genuine stylistic transformation.

Second, and more importantly: the AI cannot tell you which of the seven versions is best. It produced all seven with equal confidence and no indication of hierarchy, preference, or self-evaluation. That judgment — which version is most faithful to its claimed model, which is most interesting as a piece of writing, which is most alive — was performed entirely by the human reader. Only a native French speaker who has read Flaubert, Proust, Hugo, Balzac, and Maupassant with attention can make these discriminations. The system has no access to them.

Third, and most striking of all: the original paragraph is in several respects more interesting than most of the versions the system produced. *"Je porte cette naissance comme une sorte de titre honorifique"* has something that the seven versions largely lose: a specific authorial voice, a personal and slightly ironic self-awareness that belongs to the writer and to no one else. The AI's versions are more generically "literary" — more recognizably consistent with established stylistic conventions — but they are less distinctively *mine*. The system optimizes for the reproduction of recognizable stylistic signatures. It cannot optimize for what makes a voice original, because originality of voice is precisely what no statistical model can define or target.

A final methodological note: the exact version of ChatGPT used for this experiment was not recorded at the time. This is itself a point worth acknowledging. AI systems are updated continuously, and results obtained with one version may differ — sometimes significantly — from those obtained with another. Strict reproducibility, in the scientific sense, is therefore impossible for this kind of experiment. This is one of several reasons why the human evaluator's role remains not merely useful but structurally irreplaceable: the human brings a consistent and accountable critical perspective that the system, in whatever version, cannot supply.

**Experiment 3: Algorithm-guided music composition.** In a collaborative research project conducted at the PRISM laboratory in Marseille, a composer was asked to compose music for an experimental film from the 1930s — but without being allowed to see the film. Instead, he was given only a curve of the film's luminosity, extracted automatically by an algorithm: a simple numerical index rising and falling with the brightness of the image over time [Ariani et al. 2023]. The constraint was Oulipo-like in spirit: the algorithm provided a structure, an instruction, a displacement from the composer's habitual practice. The result was, to the research team's genuine surprise, a piece of music that cohered remarkably well with the film's imagery. More importantly, the composer himself reported that he had composed in a way he never had before — that the algorithmic constraint had opened a path he would not otherwise have taken. Here the algorithm did not compose the music. But it created the conditions under which the composer's creativity found a new expression. This is, I would argue, one of the most genuinely interesting forms of human-AI creative collaboration: not the AI as generator, but the AI as productive constraint.

**Experiment 4: Image generation, iterative dialogue, and the co-evolution of creative intention.** A fourth and analytically distinct mode of human-AI creative collaboration emerges when the AI functions as what might be called a **technical prosthetic** — a tool that enables the realization of a creative vision that the human has the intention and judgment to pursue, but not the technical means to execute manually. I have been working with the AI image generation platform Ideogram to produce a series of visual works. In each case, three conditions are present at the outset: I know what I want before I begin; I have the aesthetic expertise to evaluate whether a result achieves the intended effect; and I lack the drawing, painting, or rendering skills that would be required to produce the image by hand. The tool fills the gap between creative vision and technical execution — without filling the creative gap itself.

What makes this experiment particularly revealing, however, is the complexity of the iterative process through which each image is produced. Ideogram operates through a feature it calls the **magic prompt**: when a user submits an original prompt, the system does not generate the image directly from it, but first generates an expanded, technically elaborated interpretation — the magic prompt — which translates the user's compressed creative specification into the detailed descriptive language that image generation systems respond to most effectively. The image is generated from this magic prompt, not from the original.

In my practice, I found that I was more often refining the magic prompt than the original prompt — but not by evaluating the magic prompt in isolation. The refinement was always performed in response to two things simultaneously: the magic prompt as a text, and the image it had produced as a visual object. The image makes the AI's interpretation of my intention **visible** — it is a diagnostic tool as much as a result. Reading the image and the magic prompt together, I could identify precisely where the AI's interpretation had diverged from what I intended, and refine the magic prompt to close that gap.

But this is only half of what the iterative process involves — and the less interesting half. The more important observation is this: **the intended result is not fixed**. The original prompt encodes an intention, but that intention is not a final blueprint. As the process unfolds — as images are generated, evaluated, and refined — what becomes possible becomes visible. And what becomes visible can redirect the intention itself. A generated image may fall short of what I wanted; but it may simultaneously reveal something I had not considered, open a direction I had not foreseen, make possible something better than my original intention. In those cases, the appropriate response is not to correct the image back toward the original plan, but to follow the new possibility — to allow the process to modify the intention.

This dynamic — which might be called **creative co-evolution** — is one of the most distinctively human aspects of the entire workflow. The creative intention and the generated image co-evolve through iteration: neither is fixed; both change in response to the other. The human holds and revises the intention; the AI generates the image; the image feeds back into the intention; the revised intention generates new prompts; and so on. Donald Schön described something analogous in his account of reflective professional practice as **"reflection-in-action"** — the practitioner's capacity to revise their understanding and intentions in the middle of practice, in response to what the practice reveals [Schön 1983]. That capacity is precisely what is exercised here, at every iteration of the loop.

It is also, crucially, what AI systems cannot do. An AI system generates outputs based on its training and its prompt. It does not revise its intention based on what becomes possible — because it has no intention to revise, no capacity to recognize that a generated image opens a possibility worth following, no ability to distinguish between a result that falls short of a plan and a result that exceeds it in an unexpected direction. The human's capacity to allow the process to modify the intended result is not a loss of creative control; it is one of the most genuinely creative acts in the entire workflow. It is, in a precise sense, the human equivalent of what the paper elsewhere describes as the openness to accident and the unexpected — but here exercised not in a single unrepeatable moment, but continuously, deliberately, and iteratively throughout the process.

The images in Figure 4, shown alongside their original prompts and the magic prompts from which they were ultimately generated — with the human's refinements indicated — make this process legible. The gap between original prompt and magic prompt shows where the AI interpreted the intention. The refinements show where the human corrected that interpretation. And the relationship between successive versions, where the direction shifted, shows where the process itself redirected the intention. Taken together, they are not merely illustrations of the paper's argument. They are a record of it.

The skill required to operate this loop well is worth naming. It is not simply artistic expertise in the domain, nor technical expertise in the tool, but a new hybrid competency — **prompt literacy** — that combines the capacity to specify creative intentions in compressed language, to read the AI's interpretation of those intentions as both text and image simultaneously, to detect the gap between intention and result, and above all to remain open to the possibility that the result, even when it diverges from the intention, may be showing something worth following. Like all creative competencies, it can be developed, refined, and exercised well or poorly. And like all creative competencies, it is irreducibly human.

The first image in Figure 4 — created to illustrate the Germinals interdisciplinary seminar at Chapman University — is an instructive case of a simpler sub-mode of this process. The concept to be visualized was clear and specific: germinality, the capacity of an idea to travel from one discipline to another, to take root and grow in new soil. The original prompt encoded this intention with precision. The magic prompt generated by Ideogram expanded it faithfully, adding visual and technical specificity — colored roots representing distinct disciplines, seeds dispersing from branches and germinating below — without distorting the underlying concept. The prompt was submitted twice, each time generating four images; the image retained was selected from the second batch of four, with no further refinement of the magic prompt.

What this example illustrates is a different concentration of creative judgment from the iterative refinement cases: here, the creativity is located primarily in two acts — the specification of the original prompt, which required translating a philosophical concept into a visual brief with sufficient clarity that the AI's interpretation remained faithful; and the selection among eight generated options, which required recognizing, among outputs the system produced with equal confidence and no preference, the one that most effectively realized the intention. Neither act is available to the AI. The system cannot assess the clarity of its own interpretation of a concept it does not understand; and it cannot evaluate which of its eight outputs best achieves a goal it has no means of holding. Both judgments are irreducibly human — and both are, in the precise sense the paper has been developing, creative.

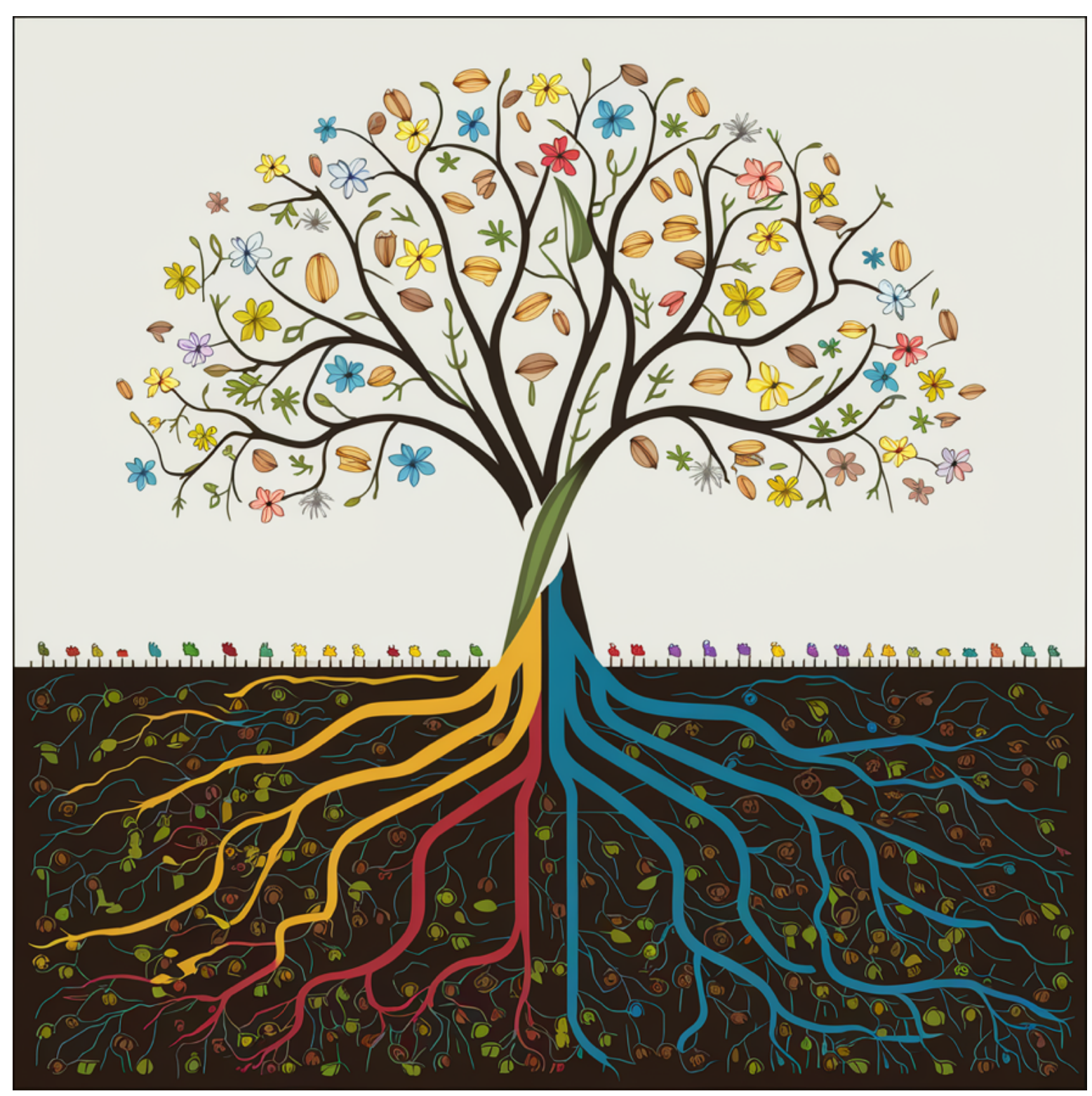

**Fig. 4a** — *Germinality* (2024). Created for the Germinals interdisciplinary seminar, Chapman University.

**Original prompt**: *"A visual to represent the notion of germinality in research, that is, the ability of an idea to germinate from one field to another, from one discipline to another. The visual should be simple, essential, and yet convey this idea of germinality in science unequivocally."*

**Magic prompt** (generated by Ideogram): *"A visually striking representation of germinality in research, featuring a stylized tree. The tree's roots extend into various academic disciplines, each represented by a different color and shape. The branches of the tree bloom with flowers that cross-pollinate, spreading their seeds to different fields. The seeds sprout in the soil of various disciplines, symbolizing the germination and growth of ideas. The overall design is minimalistic, yet effectively conveys the concept of cross-pollination and the interconnectedness of knowledge."*

**Process**: Two submissions of the same prompt, each generating four images (eight total). The image retained was selected from the second batch. No refinement of the magic prompt was required; the creative act consisted in the specification of the original prompt and the selection among multiple outputs. The AI generated; the human conceived and chose.

The second and third images in Figure 4 — *Hope for the Planet I* and *Hope for the Planet II* — illustrate a more complex sub-mode of the iterative process, and introduce a new technical element: the **seed image**. The original prompt in this case was sparse and intentionally open: *"abstract, minimalist, watercolor, transparency, hope for the planet."* The magic prompt generated by Ideogram expanded this into a fuller visual brief, specifying a central sun-like form, radiating energy, surrounding shapes representing the interconnectedness of living things, and an overall uplifting, optimistic quality. Four images were generated; the first was selected and retained as *Hope for the Planet I*.

This first image was then fed back into Ideogram as a visual seed for a second generation, alongside the original magic prompt. The system generated a new magic prompt from this combined input — and in doing so introduced an element that had not been present in the original specification: *"Jackson Pollock-inspired splatters of paint add depth and texture to the composition."* The second generation produced *Hope for the Planet II* — an image that preserved the visual DNA of the first (the central sun, the radiating lines, the watercolor quality, the color palette) while pushing it toward greater dynamism, urgency, and energy. The Pollock reference, introduced by the AI without being requested, was accepted. The result was followed rather than corrected.

This is the clearest illustration, across all four experiments, of what the paper has called the **generative branch** of the human-AI creative loop. The AI's reinterpretation of the original intention introduced a possibility — the explosive quality, the splatters, the sense of hope under pressure rather than hope as stillness — that the original intention had not contained. The human recognized this possibility as valuable and followed it. The intended result was not corrected back toward its original form; it was expanded by what the process revealed. The two images together form a diptych that would not have existed had the creative intention been fixed in advance: *Hope for the Planet I* is hope as a quiet, interior state; *Hope for the Planet II* is hope as struggle, turbulence, and energy. The diptych is richer than either intention — the original or the AI-modified one — would have produced alone.

A final and deeper observation must be made here, one that the preceding analysis of individual experiments does not fully capture. The process of working with an AI image generation system across multiple sessions — specifying prompts, evaluating magic prompts, selecting among outputs, following generative surprises, correcting interpretive failures, feeding images back as seeds — is not only productive. It is **formative**. Each collaboration develops the human artist's creative capacities: their prompt literacy, their ability to recognize what works and what does not, their understanding of what is possible, their clarity about their own aesthetic intentions. The collaboration is not just a means of producing images; it is a practice that shapes the practitioner. As Marshall McLuhan observed, "we shape our tools, and thereafter our tools shape us" [McLuhan 1964]. The shaping here is not passive or mechanical — it is creative, iterative, and cumulative. And it belongs entirely to the human side of the collaboration. The AI, in whatever version and however many sessions, learns nothing from the exchange. Its capacities do not develop through the collaboration. The human's do. This asymmetry — the human grows, the machine does not — is perhaps the most fundamental of all the distinctions this paper has sought to draw.

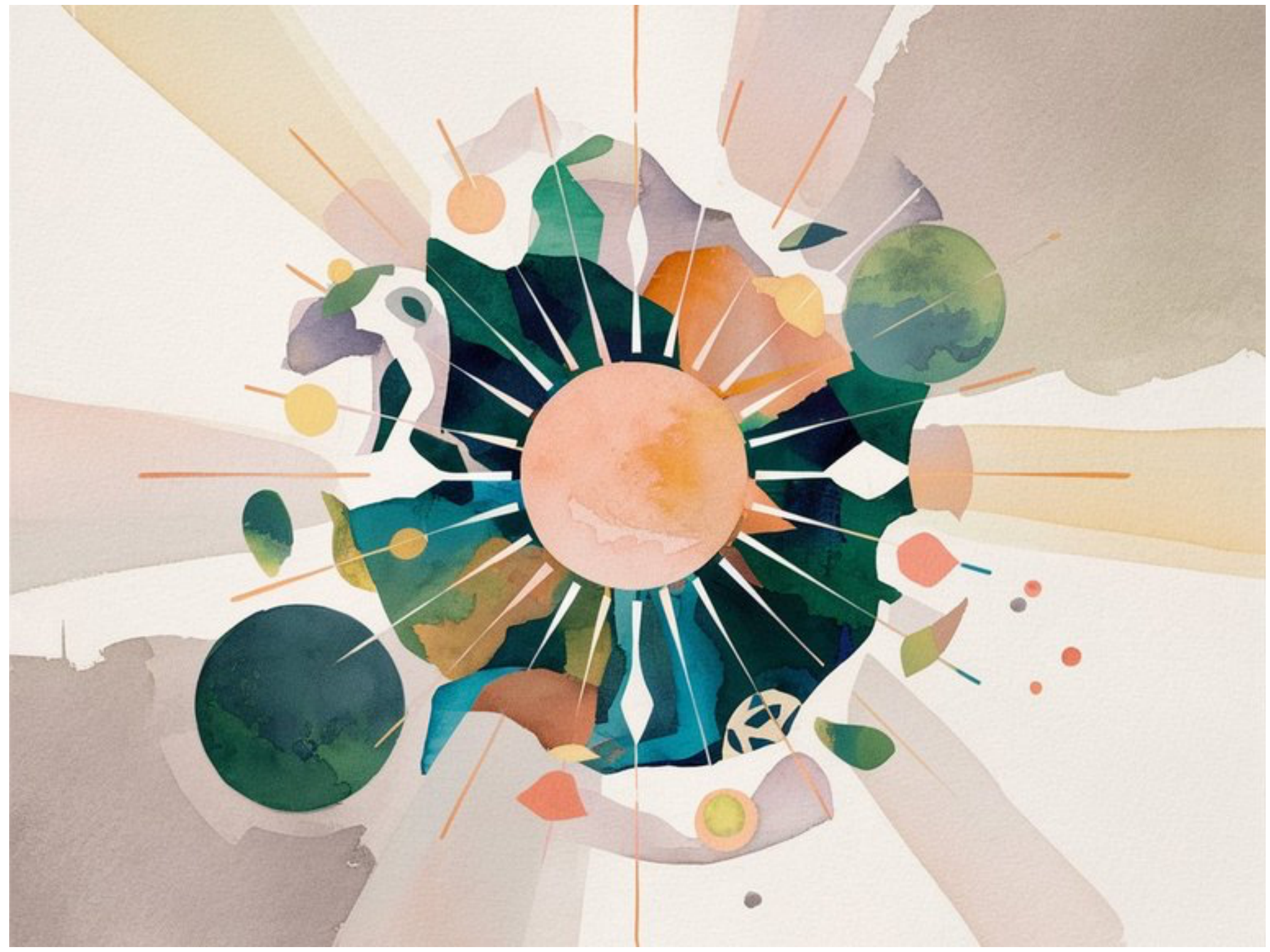

**Fig. 4b** — *Hope for the Planet I* (2024).

**Original prompt**: *"Abstract. Minimalist. Watercolor. Transparency. Hope for the planet."*

**Magic prompt 1** (generated by Ideogram): *"A mesmerizing abstract watercolor painting with minimalist elements, featuring a blend of transparencies and vibrant colors. The composition represents hope for the planet, with a central sun-like shape emitting rays of light and energy. Surrounding this central figure, there are various shapes and forms that could represent the interconnectedness of all living things on Earth. The overall effect is uplifting and full of optimism, inspiring viewers to cherish and protect our planet."*

**Process**: Four images generated from the first submission; this image was selected and retained. It subsequently served as a visual seed for the second generation.

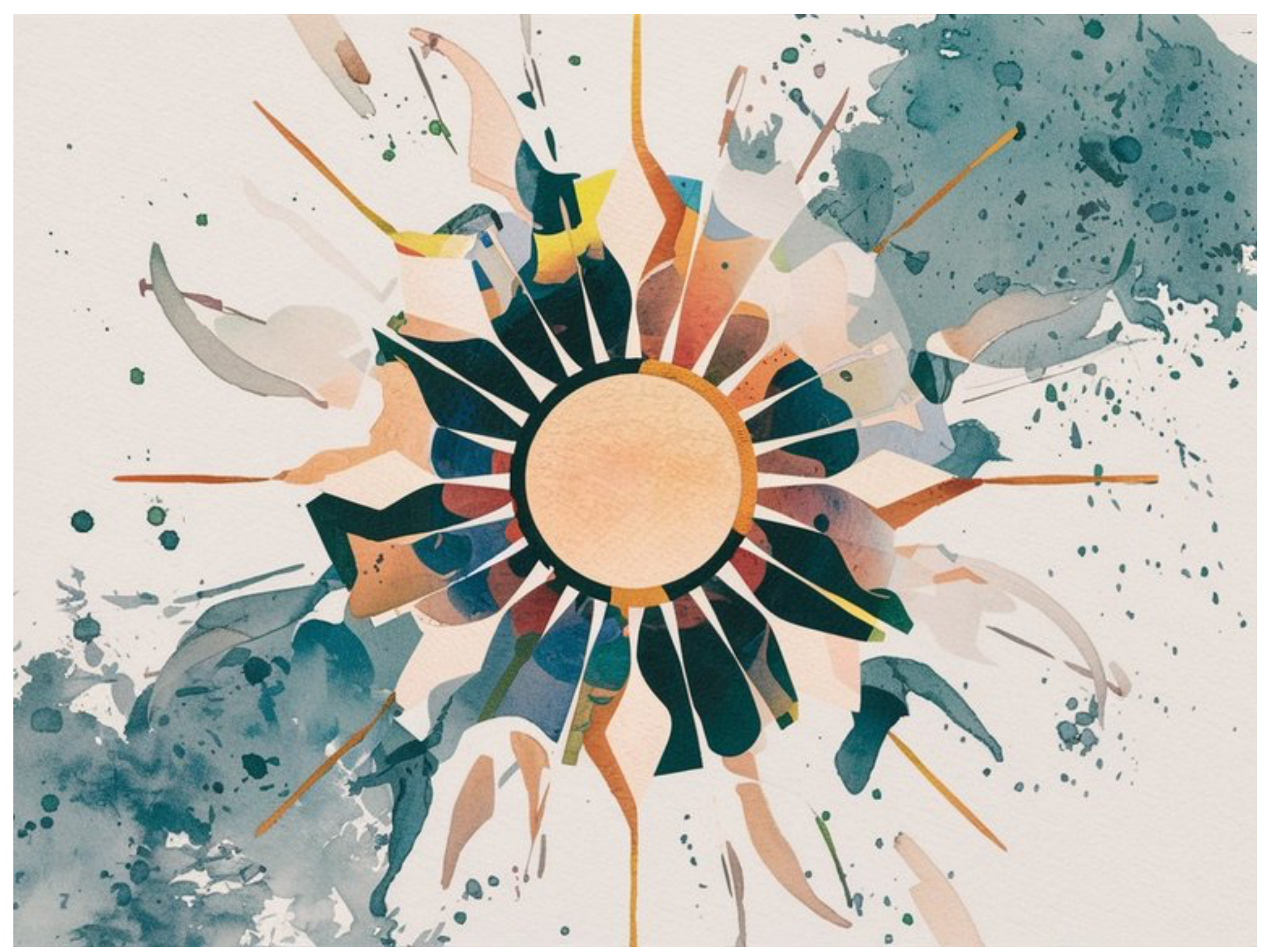

**Fig. 4c** — *Hope for the Planet II* (2024).

**Seed image**: *Hope for the Planet I* (Fig. 4b), fed back into Ideogram as a visual input alongside Magic Prompt 1.

**Magic prompt 2** (generated by Ideogram from seed image + Magic Prompt 1): *"A captivating abstract watercolor painting, expertly blending minimalist elements with mesmerizing transparencies and vibrant colors. The central focus is a sun-like shape, radiating rays of light and energy, symbolizing hope for our planet. Surrounding this core element are intricate shapes and forms, representing the interconnectedness of all living beings on Earth. Jackson Pollock-inspired splatters of paint add depth and texture to the composition. The overall effect is uplifting and brimming with optimism, motivating viewers to cherish and protect our planet."*

**Note**: The Pollock reference was introduced by Ideogram, not specified in the original prompt. It was accepted and followed. This image illustrates the generative branch of the human-AI creative loop: the AI's interpretation expanded the human's original intention rather than simply executing it.

The watercolor that composed itself through humidity over fifteen years in an attic was a pure accident — unrepeatable, unplanned, and recognized only in retrospect. The iterative image generation process described here is something different: a structured and deliberate openness to the unexpected, exercised in real time, within a collaborative loop. Both are instances of the same fundamental creative disposition — the willingness to be redirected by what one cannot foresee. That disposition is what no AI system, however sophisticated, currently possesses.

## 6.2 The Conditions for a Productive Collaboration

These four experiments suggest that AI can be a productive creative collaborator under specific conditions. The most important condition is that the human partner must have a high level of domain expertise — not merely technical expertise in using the AI system, but genuine artistic expertise in the domain in question. The value of the collaboration is entirely dependent on the human's ability to evaluate the AI's outputs and identify what is interesting, surprising, or useful among them.

This has an implication that is often overlooked in public discussions of AI creativity: the quality of the human-AI collaboration is, paradoxically, a measure of the *human's* creativity, not the AI's. The AI provides material; the human provides judgment. And it is the quality of the judgment — the capacity to recognize value in unexpected places — that determines whether the collaboration produces something genuinely creative.

AI systems can serve productively as a sounding board and brainstorming partner in the early stages of a creative process. They can be useful in testing the structural properties of an emerging work — identifying gaps, inconsistencies, or unexplored possibilities. They can provide rapid access to relevant precedents, styles, and forms from across a vast corpus. All of these contributions have real value in a creative process.

There is, however, a further and perhaps more fundamental mechanism of productive collaboration that deserves emphasis: what I have elsewhere called *déplacement*, or displacement [Magrin-Chagnolleau, in Press]. The idea draws on the tradition of the Oulipo — the French literary movement founded in the 1960s by Raymond Queneau, Georges Perec, Italo Calvino, and others, which explored how formal constraints could stimulate rather than restrict creativity [Oulipo 1988]. The most celebrated example is Perec's *La Disparition* (1969), an entire novel written in French without the letter *e* — the most common letter in the language [Perec 1969]. The constraint does not limit creativity; it forces it to find another path. An AI system, used well, can serve an analogous function: not as a generator of finished creative products, but as a device that forces the human artist to do something different from what they would otherwise have done — to step outside habitual practice, to be surprised by the result of following an unexpected instruction. In this sense, the most creative use of an AI system in artistic practice may not be to ask it to create, but to ask it to *constrain* or *displace* — to become, in effect, a generator of productive accidents that the human artist then navigates with their full creative intelligence. Experiment 3 above is a precise illustration of this principle: the algorithm's luminosity curve did not compose the music, but it made the composer compose differently.

## 6.3 The Danger of the Average

There is, however, a significant danger in over-reliance on AI collaboration: the gravitational pull toward the average. Because AI systems are trained to produce statistically probable outputs, their tendency is to generate content that is consistent with the most common patterns in their training data. In creative terms, this means a tendency toward the familiar, the conventional, the expected, the *cliché*. The outputs are rarely offensive, rarely embarrassing — but they are also rarely surprising in the deepest sense.

An artist who uses an AI system as a creative partner without maintaining critical distance risks having their creative decisions subtly shaped by this gravitational pull. The AI's suggestions will tend to be good — competent, well-formed, consistent with existing conventions — and this very competence can make them difficult to resist. The result is a creative work that is smooth, polished, and slightly less original than it might otherwise have been.

This danger is real, and it reinforces the importance of the condition established above: only an artist with genuine domain expertise can recognize the AI's tendency toward the average and resist it — can use the AI's outputs as a foil, a starting point, or a provocation, rather than as a destination. There is also a responsibility that falls on researchers and practitioners working in this field: to resist the instrumentalization of art by cultural and creative industries that have a strong economic interest in overstating AI's creative capabilities, and to be vigilant about what is claimed, written, and published on this subject [Magrin-Chagnolleau, in Press].

## (a) FOUR EXPERIMENTS: HUMAN–AI CREATIVE COLLABORATION

### 1. Haiku composition
(Autumn 2025, ChatGPT)

AI generates; human evaluates for novelty, surprise and value.

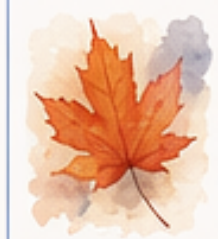

*Crimson leaves descend,*
*Whispering secrets of fall,*
*Earth's warm, final blend.*

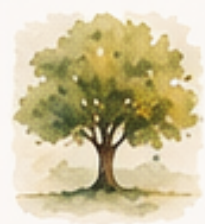

*Consciousness in trees,*
*Creativity in leaves,*
*Free will in the breeze.*

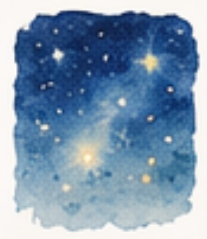

*Stars in endless night,*
*Universe in boundless flight,*
*Minds expand in light.*

- 60 haikus in total (6 series).
- Formal competence: high.
- Genuine surprise/novelty: rare.
- Self-evaluation by AI: absent.

### 2. Autobiography in the style of 5 French literary masters
(Spring 2026, ChatGPT)

AI combines styles; human judges fidelity, interest and value.

**Original paragraph (excerpt)**

*Je suis né à Paris. En fait, je suis né à Montmartre, ce qui aujourd'hui n'est plus possible car la clinique où je suis né, la clinique Ornano rue Championnet, n'existe plus… Je ne sais pas pourquoi j'y suis né. Mais je porte cette naissance comme une sorte de titre honorifique.*

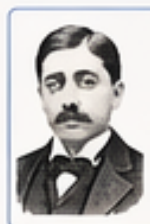

**Proust**

*ayant disparu depuis longtemps, comme tant de lieux qui furent autrefois les témoins silencieux de nos premiers instants…*

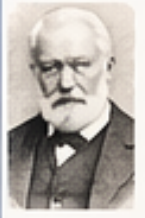

**Hugo**

*Paris ! Paris, la ville lumière, la cité des révolutions et des rêves ! […] Et cette naissance est pour moi comme un sceau, une marque, une étoile…*

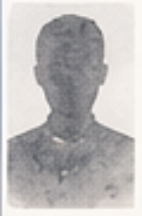

**Synthèse moderne (v.7)**

*Pour naître là-haut, il faudrait désormais le faire chez soi — et qui choisit encore cela ? … Un trajet à rebours, comme si je remontais vers le cœur de la ville…*

- Excellente imitation stylistique.
- Aucune capacité d'auto-évaluation.
- Voix personnelle ≠ reproduite.

### 3. Algorithm-guided music composition
(PRISM Lab, 2023)

Algorithm provides constraint; human composes and evaluates.

Luminosity curve of an abstract film (*Rhythmus 21*, Hans Richter, 1921)
Excerpt: 1m30s

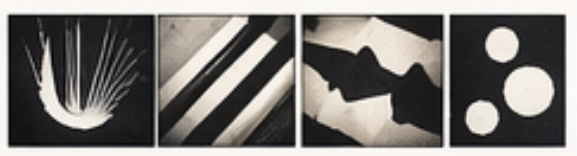

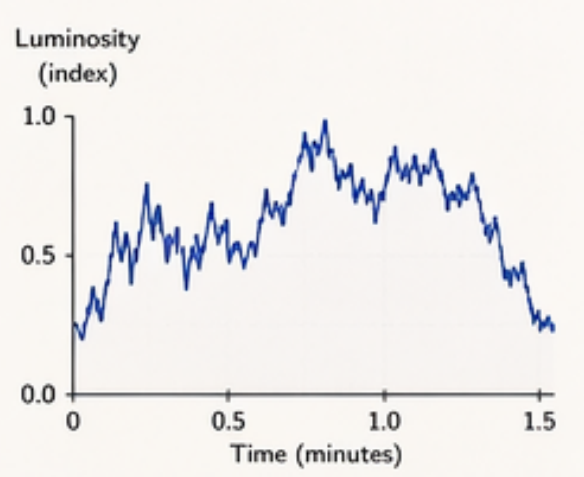

Composed instrumental music from luminosity parameters

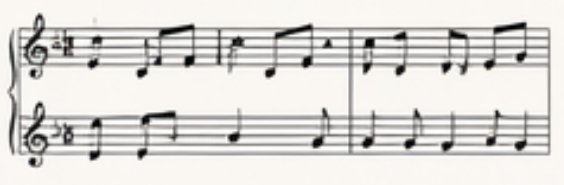

- Surprising synchronicity with the film.
- Algorithm = productive constraint.
- Music creativity = human.

### 4. Image generation with Ideogram
(2024)

Magic prompt, iterative dialogue, creative co-evolution.

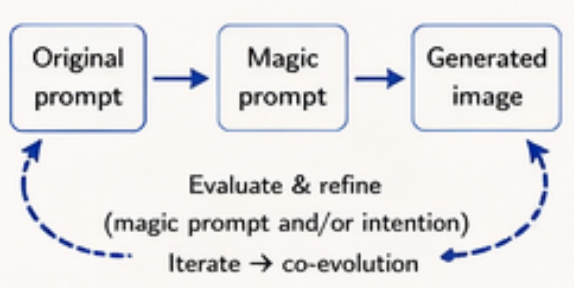

Image 1 (seed) — Image 2 (evolved)

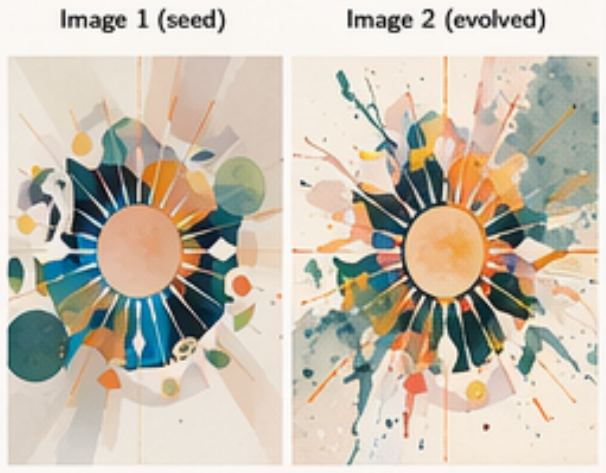

- Prompt literacy + human judgment.
- Intention can be redirected by the process.
- Human grows; AI does not.

## (b) FOUR STRUCTURAL LIMITATIONS OF AI SYSTEMS WITH RESPECT TO BODEN'S THREE PROPERTIES (NOVELTY, SURPRISE, VALUE)

### 4.1 Data-bound: No Historical Novelty

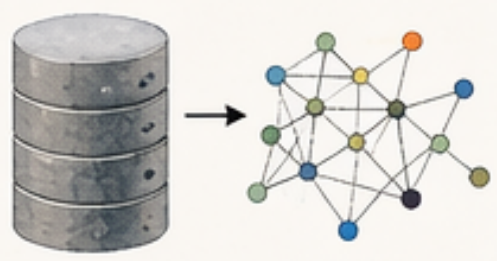

AI systems rely on large datasets of existing human-generated content. They cannot produce what has no precedent in their training data.

→ No historical novelty.

### 4.2 Statistical Convergence: No Genuine Surprise

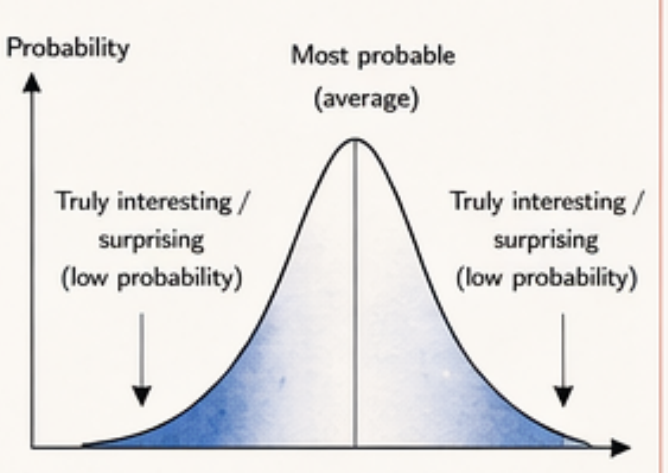

Outputs are drawn from the highest-probability region. Creativity often comes from derailment, not average.

→ No genuine surprise.

### 4.3 Evaluation Problem: No Access to Value

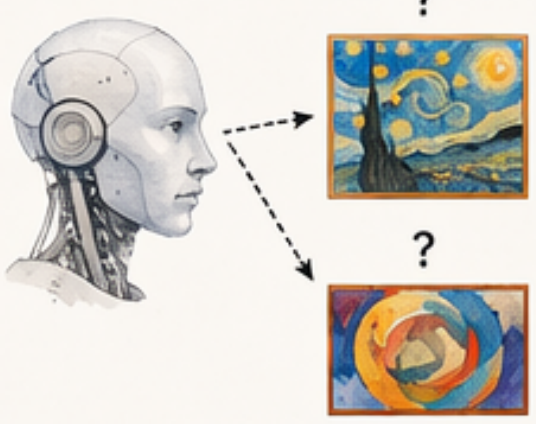

AI can compare to past evaluations but cannot judge aesthetic, emotional, or intellectual value from within. Value requires a subject.

→ No evaluation of value.

### 4.4 No Accident, No Serendipity, No Subjectivity

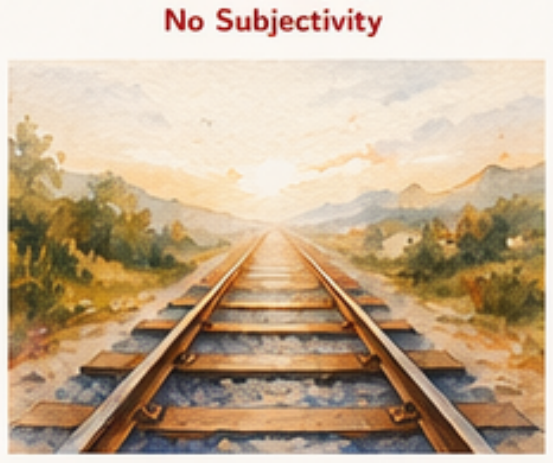

AI follows the track of probability. No genuine accident. No capacity to notice, welcome, and use the unexpected (requires a subject).

→ No accident, no serendipity, no subjectivity.

## (c) FOUR DIMENSIONS OF HUMAN CREATIVITY STRUCTURALLY ABSENT FROM CURRENT AI SYSTEMS

### 1. Openness to the Unexpected (Accident)

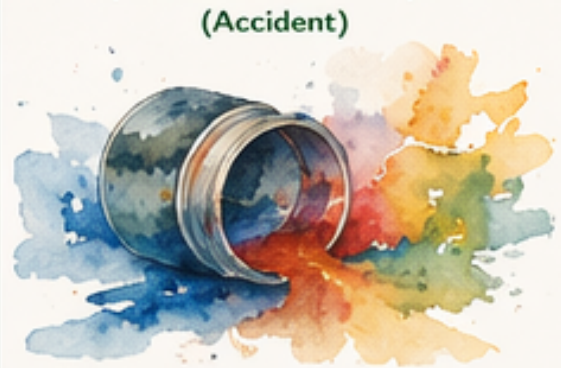

Creative breakthroughs often arise from accidents, mistakes, and the unforeseen. A source of real novelty.

### 2. Subjective Reception of Chance (Bergson / Valéry)

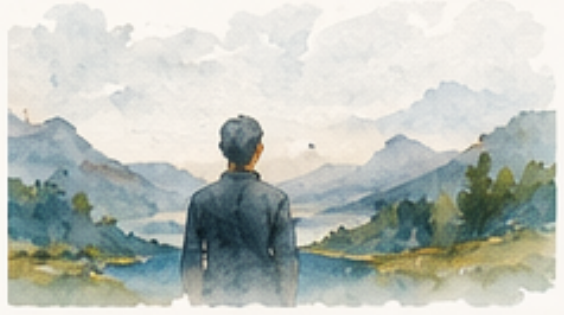

Chance exists only for a subject who perceives, recognizes, and welcomes it. No subject → no chance.

### 3. Provocation and Readiness (Camus / Rolland)

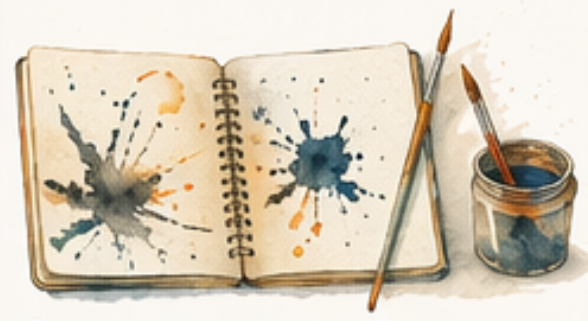

Chance must be provoked. The creator cultivates a state of attention, openness, and readiness to follow what appears.

### 4. Transformational Capacity

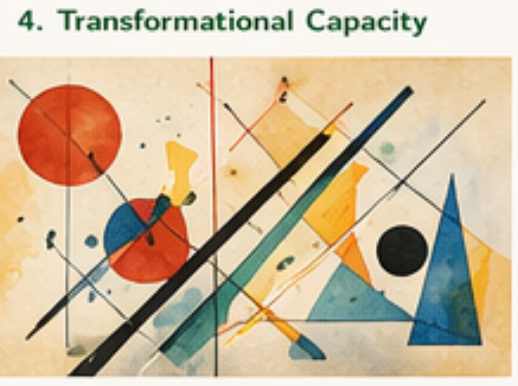

Transformational creativity restructures the conceptual space itself. AI cannot change the axioms of its conceptual space.

**Figure 5.** Synthetic overview of the paper's argument generated by AI. **(a)** Four empirical experiments illustrating different modes of human-AI creative collaboration (Section 6.1): (1) haiku composition — AI as ideation partner, human evaluates (autumn 2025, ChatGPT); (2) autobiography in the style of five French literary masters — AI as stylistic combinatorial engine, human judges (spring 2026, ChatGPT); (3) algorithm-guided music composition — algorithm as productive constraint, using a 1m30s excerpt of *Rhythmus 21* (Hans Richter, 1921) [Ariani et al. 2023]; (4) image generation with Ideogram — AI as technical prosthetic, magic prompt, creative co-evolution (2024). **(b)** Four structural limitations that prevent AI systems from satisfying Boden's three properties of creativity in the strongest sense (Sections 4.1–4.4). **(c)** Four dimensions of human creativity — phenomenological and transformational — that are structurally absent from current AI systems (Sections 4.4 and 5). All images are either public domain, belonging to the author, or generated specifically for this article; selection and integration by the author.

# 7. Conclusion

The question "can an AI system be creative?" does not admit of a simple yes or no. It requires, first, a precise definition of creativity. Boden's framework — distinguishing three properties and three types of creative process — provides the best available precision. With that framework in place, the answer becomes graduated and specific.

In the domain of **combinatorial creativity**, AI systems have genuine capabilities. They can make connections across vast corpora that no individual human could, and these connections can be generative for a human creative process. But they cannot evaluate the value of the combinations they produce, and their structural tendency is toward the statistically probable rather than the genuinely surprising.

In the domain of **exploratory creativity**, AI systems can traverse an existing conceptual space rapidly and extensively, but they cannot recognize the edges of that space or understand what it would mean to approach and exceed them. Their exploration is bounded by their training data, and they have no vantage point from which to see beyond that boundary, nor any capacity to evaluate the relative interest of different positions within the space they are exploring.

In the domain of **transformational creativity**, AI systems are, at present and by structural necessity, incapable. Transformational creativity requires stepping outside an existing conceptual framework and restructuring it — an act that is, by definition, impossible for a system whose conceptual space is defined and bounded by its training data.

The deepest reason for these limitations is not purely technical but phenomenological: genuine creativity, at its most powerful, is not merely the production of novelty from existing elements. It is a process that is *open to being redirected by what it cannot foresee*

— by accident, by the unexpected, by the serendipitous failure that opens a new path. And it requires a subject who is in a state of readiness to receive and recognize such events — who can, in Camus's words, give opportunities to chance, and in Rolland's, know how to use it when it arrives. Both dimensions — the accident itself and the subject who receives it — are structurally absent from current AI systems.

None of this means that AI systems are useless for creative work. On the contrary: in the hands of a human expert with the skill to guide the collaboration and the judgment to evaluate its outputs, an AI system can be a powerful creative partner — as brainstorming tool, as combinatorial engine, and above all as a device for productive displacement, forcing the human artist to do something different from what they would otherwise have done. But the locus of creativity in such a collaboration is the human, not the machine.

The paper you have just read is itself a case in point. The thoughts are mine, shaped by years of practice in engineering, the arts, and philosophy. The AI contributed to organization, translation, and literature review. But the thesis — the claim that AI systems cannot be creative in the strongest sense of the word — is mine. The evaluation of whether that claim has value is yours. And the practice of working with AI systems — across all four experiments described here — has developed capacities that are mine alone, and that no amount of AI generation could have produced without the human intelligence that directed, evaluated, and learned from every step of the process. We shape our tools, and thereafter our tools shape us [McLuhan 1964]. What they cannot do, as yet, is shape themselves.